\documentclass[conference]{IEEEtran}
\IEEEoverridecommandlockouts
\usepackage{cite}
\usepackage{amsmath,amssymb,amsfonts}
\usepackage{algorithmic}
\usepackage{graphicx}
\usepackage{textcomp}
\usepackage{xcolor}
\usepackage{arydshln}

\usepackage{times}
\usepackage{epsfig}
\usepackage{diagbox}
\usepackage[linesnumbered,algoruled,boxed,lined]{algorithm2e}
\usepackage{subcaption}
\usepackage{multirow,tabularx}
\usepackage[normalem]{ulem}
\usepackage{array}
\usepackage{colortbl}
\usepackage{tabularx}

\makeatletter
\def\BState{\State\hskip-\ALG@thistlm}
\makeatother

\newcommand\blfootnote[1]{%
  \begingroup
  \renewcommand\thefootnote{}\footnote{#1}%
  \addtocounter{footnote}{-1}%
  \endgroup
}

\def\ttabular{%
\hbox\bgroup
\let\\\cr
\def\rulea{\ifnum\rowc=\@ne \hrule height 1.3pt \fi}
\def\ruleb{
\ifnum\rowc=1\hrule height 1.3pt \else
\ifnum\rowc=6\hrule height \heavyrulewidth 
   \else \hrule height \lightrulewidth\fi\fi}
\valign\bgroup
\global\rowc\@ne
\rulea
\hbox to 10em{\strut \hfill##\hfill}%
\ruleb
&&%
\global\advance\rowc\@ne
\hbox to 10em{\strut\hfill##\hfill}%
\ruleb
\cr}
\def\endttabular{%
\crcr\egroup\egroup}

\begin{document}

\title{Variational Clustering: Leveraging Variational Autoencoders for Image Clustering}

\author{\IEEEauthorblockN{Vignesh Prasad*}
\IEEEauthorblockA{\textit{TU Darmstadt}\\
Germany \\
vignesh.prasad@tu-darmstadt.de}
\and
\IEEEauthorblockN{Dipanjan Das*}
\IEEEauthorblockA{\textit{Embedded Systems and Robotics} \\
\textit{TCS Innovation Labs}, Kolkata, India\\
dipanjan.da@tcs.com}
\and
\IEEEauthorblockN{Brojeshwar Bhowmick}
\IEEEauthorblockA{\textit{Embedded Systems and Robotics} \\
\textit{TCS Innovation Labs}, Kolkata, India\\
b.bhowmick@tcs.com}
}

\maketitle
\blfootnote{This work was done when Vignesh Prasad worked at TCS Innovation Labs. * - Equal Contribution}

\begin{abstract}
Recent advances in deep learning have shown their ability to learn strong feature representations for images. The task of image clustering naturally requires good feature representations to capture the distribution of the data and subsequently differentiate data points from one another. Often these two aspects are dealt with independently and thus traditional feature learning alone does not suffice in partitioning the data meaningfully. Variational Autoencoders (VAEs) naturally lend themselves to learning data distributions in a latent space. Since we wish to efficiently discriminate between different clusters in the data, we propose a method based on VAEs where we use a Gaussian Mixture prior to help cluster the images accurately. We jointly learn the parameters of both the prior and the posterior distributions. Our method represents a true Gaussian Mixture VAE. This way, our method simultaneously learns a prior that captures the latent distribution of the images and a posterior to help discriminate well between data points. We also propose a novel reparametrization of the latent space consisting of a mixture of discrete and continuous variables. One key takeaway is that our method generalizes better across different datasets without using any pre-training or learnt models, unlike existing methods, allowing it to be trained from scratch in an end-to-end manner. We verify our efficacy and generalizability experimentally by achieving state-of-the-art results among unsupervised methods on a variety of datasets. To the best of our knowledge, we are the first to pursue image clustering using VAEs in a purely unsupervised manner on real image datasets.
\end{abstract}

\begin{IEEEkeywords}
Unsupervised Learning, Clustering, Variational Inference
\end{IEEEkeywords}

\section{Introduction}
\begin{figure*}[h!]
	\centering
	\begin{tabular}{cccc}
	\centering
	\begin{subfigure}[b]{0.2\linewidth}
    	\centering
        \includegraphics[width=0.9\linewidth]{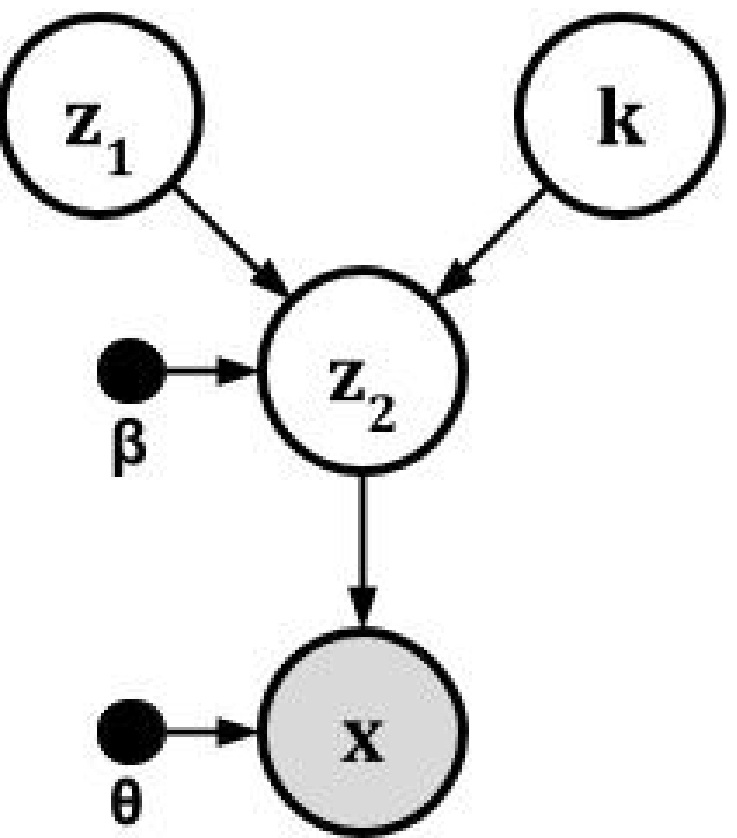}
        \caption{GMVAE Gen}
        \label{fig:gmvae-gen}
    \end{subfigure}&
    \begin{subfigure}[b]{0.2\linewidth}
    	\centering
        \includegraphics[width=0.95\linewidth]{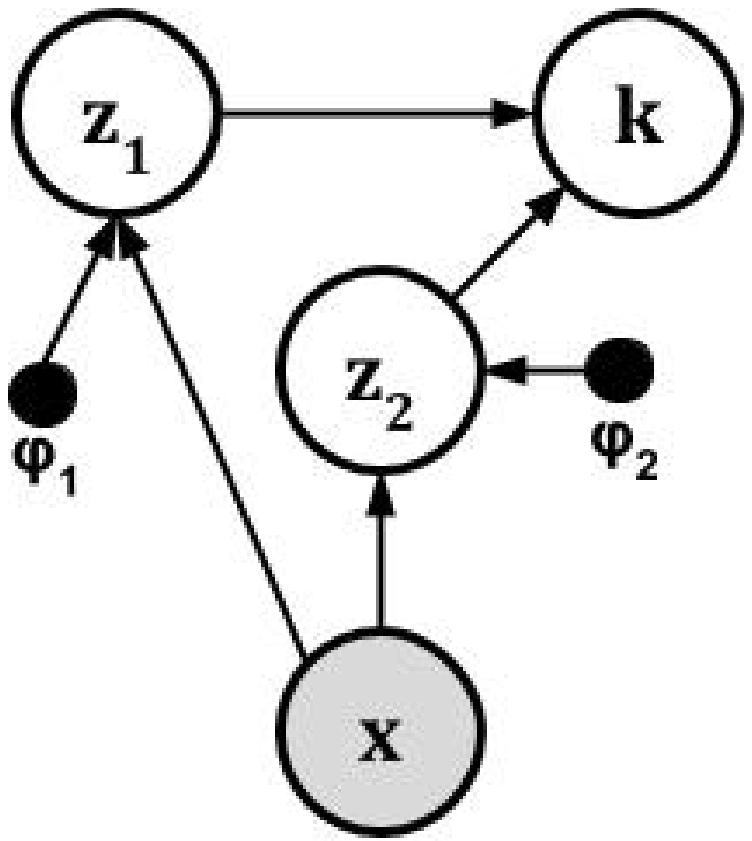}
        \caption{GMVAE Inf}
        \label{fig:gmvae-inf}
    \end{subfigure}& 
    \begin{subfigure}[b]{0.2\linewidth}
    	\centering
        \includegraphics[width=0.5\linewidth]{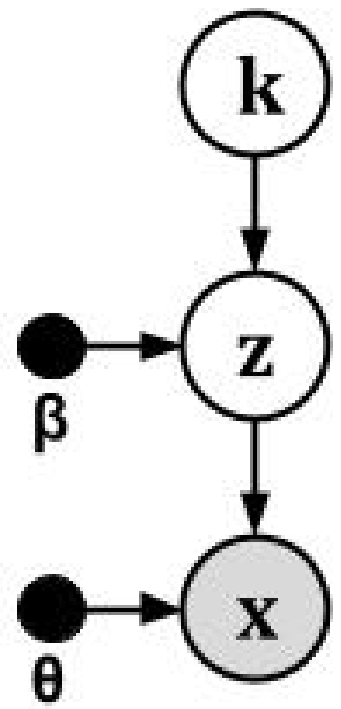}
        \caption{Ours Gen}
        \label{fig:our-gen}
    \end{subfigure}&
    \begin{subfigure}[b]{0.2\linewidth}
    	\centering
        \includegraphics[width=0.6\linewidth]{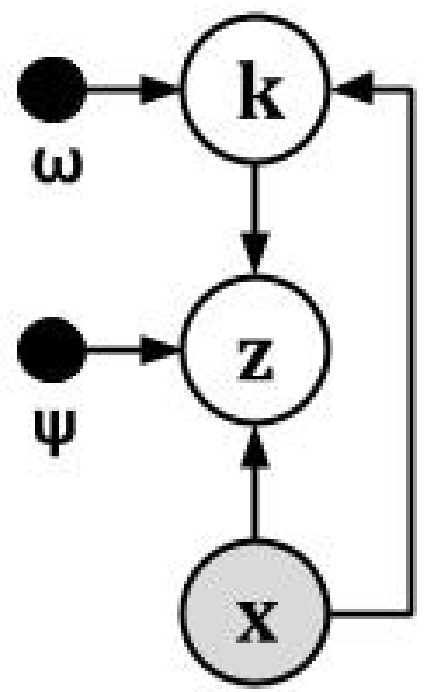}
        \caption{Ours Inf}
        \label{fig:our-inf}
    \end{subfigure}
	\end{tabular}
	\vspace{-1em}
	\caption{Graphical representation of the inference and generative models used by us and GMVAE\cite{dilokthanakul2016deep}. VaDE\cite{jiang2017variational} also has a similar generative model as ours but differ in their inference model. Grey nodes represent data nodes. White ones represent latent nodes, where $z$, $z_1$ and $z_2$ are normally distributed variables are $k$ is a discrete variable. Black nodes represent the network parameters. (Gen - Generative, Inf- Inference)}
	\label{fig:graph}
	\vspace{-2em}
\end{figure*}
Image Clustering is a fundamental, challenging and widely studied problem in machine learning \cite{ng2002spectral,ye2008discriminative,yang2010image,von2007tutorial,yang2016joint,gdalyahu2001self}. with variety of applications in image retrieval \cite{retrieval}, fast 3D reconstructions \cite{sfm} \cite{Broj2} \cite{Broj4} etc. Some classical examples are $K$-means \cite{macqueen1967some}, Gaussian Mixture Models \cite{bishop2006pattern} and Spectral clustering \cite{von2007tutorial} which are promising, but require a robust feature representation for good clustering. In recent years, Deep Learning has made huge progress in learning robust feature representations of images. These learned representations help cluster the data more accurately when used with traditional methods like $K$-means for example\cite{pmlr-v70-yang17b}. One way to use deep representations, off the shelf, is to extract the feature representation of an image from a pre-trained model and use them directly in any clustering algorithm\cite{tian2014learning}. The problem with such approaches is that they don't fully exploit the power of deep neural networks. Song \textit{et al.} \cite{song2013auto} learn a representation to accurately cluster the images in the dataset by integrating $K$-means into the bottleneck layer of an Autoencoder. This association enables the model to learn a meaningful clustering-oriented representation. \par 



With the motivation to pursue a robust and generalizable methodology in a principled way, we aim to make inferences in a latent space learned specifically for a clustering task. The idea is that it would be easier to group the data in this space, compared to an arbitrary space defined by pre-trained features. Off late, the use of generative methods for clustering has been on the rise as their expressive power helps efficiently capture, represent and recreate sampled data points. As we wish to experiment with data distributions in a latent space that can accurately represent the input data, the paradigm of Variational Autoencoders (VAEs) lends itself directly to the task at hand.\par 


We build on the ideas of GMVAE\cite{dilokthanakul2016deep} and VaDE\cite{jiang2017variational} addressing their fallacies while maintaining the underlying motivation of using a Gaussian Mixture Model as the latent space distribution. Instead of deriving the prior from a random variable, as in GMVAE, our prior is deterministic. This is similar to VaDE however, we learn the parameters for the prior and posterior jointly, unlike in VaDE which uses a pre-training phase to initialize the parameters of the prior. \par 

To illustrate the differences between our process, GVMAE\cite{dilokthanakul2016deep} and VaDE\cite{jiang2017variational} we visualize the graphical models in Fig. \ref{fig:graph}. In GMVAE, the Gaussian prior $\mathbf{z_2}$ depends on a noise variable $\mathbf{z_1}$ \& varies for a given cluster, as shown in Fig. \ref{fig:gmvae-gen}. Ours is more intuitive as it depends only on the cluster, as shown in Fig. \ref{fig:our-gen}. Secondly, GMVAE expresses the categorical posterior $q(k|\mathbf{z_1},\mathbf{z_2})$ with the prior $p_\beta(\mathbf{z_2}|\mathbf{z_1},k)$ using Bayes' rule. This applies to VaDE too, along with fixing the GMM prior during pre-training. We learn it during training, giving more flexible learning, the effectiveness of which is seen in the results in Table \ref{tab:accuracy}. This can also be seen on a toy dataset, compared to GMVAE, where our method learns a more compact cluster representation as compared to GMVAE, as shown in Fig. \ref{fig:toy_data}.

Our method is more principled as we directly learn the cluster assignment probabilities $q(k|\mathbf{z})$ instead of performing a Bayesian classification, as done in both GMVAE and VaDE. Once the cluster predictions $q(k|\mathbf{z})$ become closer to a one-hot vector, it, in turn, minimizes the loss for just a given cluster instead of over all clusters leading to a more effective clustering. This process is shown in Fig. \ref{fig:flow}.


Along with this, we propose a novel augmentation loss as a mixture of $L2$ distance between the cluster predictions and 2-Wasserstein distance\cite{vaserstein1969markov} between the predicted posterior distributions. We do so to ensure consistency in the cluster predictions and also in the predicted posteriors thereby enforcing a similarity among samples from a given cluster. 
The idea behind this is to achieve invariance to marginal variations in the data and predict an accurate mapping for both real as well as augmented data. It would be counter-intuitive to map an augmented version of an input to a different distribution


Therefore, our novelty is in modelling a GMM prior whose parameters are learnt during training, which is more flexible than fixed distributions like in other methods. This is too rigid a constraint mainly in complex datasets (CIFAR/STL). Due to our principled \& effective modelling, coupled with our novel incorporation of an augmentation loss, our accuracy is significantly better than others and is very coherent on similar datasets. We perform well even in a cross-dataset transfer.

The main contributions of this paper are:
\begin{itemize}
\item A novel unsupervised image clustering algorithm by combining VAEs with a true Gaussian Mixture prior learnt without any pre-training in an end-to-end manner.
\item More principled latent space priors that subsequently lead to a simpler inference model.
\item A novel augmentation loss making our method robust and leading to strong intra-cluster relations.
\end{itemize}
In contrast to other methods, our method performs consistently well over all datasets using only raw pixel inputs, showcasing the generalization capability of our method. To the best of our knowledge, we are the first to do so using VAEs in a purely unsupervised manner, especially on real image datasets.

\section{Related Works}
Tian \textit{et al.}\cite{tian2014learning} explore the use of deep methods in classical clustering by first learning a feature representation and subsequently performing clustering in the feature space. Nina et al.\cite{Nina_2019_ICCV} follow a similar approach by using a triplet loss between positive and negative anchor images from the dataset and perform K-means on the learnt latent space. Similarly, Das et al.\cite{das2019deep} propose a novel method for selecting images for calculating a triplet loss, which however depends on the quality of the pre-training. Yang \textit{et al.} \cite{yang2019deep} follow a similar approach for learning a latent space but follow a graph-based spectral clustering approach with the learnt representation.

Song \textit{et al.} \cite{song2013auto} take a step further by integrating $K$-means into the bottleneck of an Autoencoder. This enables the model to learn meaningful clustering-oriented representations. Deep Embedded Clustering (DEC)\cite{xie2016unsupervised} builds on a similar idea by minimizing the KL Divergence between cluster assignment probabilities and an auxiliary target distribution derived from the current predictions. However, this causes the auxiliary distribution to change as the predictions do during training. \par 

Deep Adaptive Clustering (DAC)\cite{Chang_2017_ICCV} uses a pairwise classification approach to distinguish between image pairs. It learns features that correspond directly to predicted class labels and achieve reasonable accuracy. Similarly, Ji \textit{et al.} \cite{ji2019invariant} propose a method that tries to preserve pair-wise semantic mutual information in the data. 
Deep Embedded Regularized Clustering (DEPICT) \cite{dizaji2017deep} uses a convolutional autoencoder to jointly learn representations and cluster predictions by constraining clusters to have similar no. of samples. Their performance deteriorates when data is not distributed well across clusters. To this end, RDEC\cite{pmlr-v95-tao18a} uses adversarial training to improve performance on imbalanced datasets. Along similar lines, Joint Unsupervised Learning (JULE) \cite{yang2016joint} learns representations and cluster assignments simultaneously using agglomerative clustering. Though it gets good results, their method is computationally intensive and has a heavy memory usage \cite{hsu2018cnn}. Sarfaraz \textit{et al.} \cite{sarfraz2019efficient} also develop an agglomerative clustering method but with lower overheads in computing distances by partitioning the data points more efficiently. \par 

Haeusser \textit{et al.}\cite{haeusserassociative} attempt to enforce intra-class similarity with associative constraints between images and their augmented versions to ensure similarity in predictions. However, the robustness and generalization capability of their method are questionable since they show vastly differing results on CIFAR10\cite{krizhevsky2009learning} and STL-10\cite{coates2011analysis}, both of which are real-world datasets having 9 same classes out of 10. \par

Information-Maximizing Self-Augmented Training (IMSAT) \cite{pmlr-v70-hu17b} learn a probabilistic classifier using Regularized Information Maximization (RIM) \cite{krause2010discriminative}. For feature-rich datasets (CIFAR10 and CIFAR100 \cite{krizhevsky2009learning}), they use ResNet\cite{he2016deep} features as input. Thus, it is difficult to properly gauge their performance as results without pre-trained models are unavailable.\par


Coming to Deep Generative methods, ClusterGAN \cite{mukherjee2019clustergan} is a Generative Adversarial framework for unsupervised clustering using a discrete-continuous mixed approach for clustering. Ghasedi \textit{et al.}  \cite{ghasedi2019balanced} build on this idea by introducing a self-paced learning algorithm that helps guide the learning, similar to a curriculum learning framework, to get better results. Variational Deep Embedding (VaDE) \cite{jiang2017variational} and Gaussian Mixture VAEs (GMVAE) \cite{dilokthanakul2016deep} use VAEs with GMMs simultaneously to model the inference process. Yang \textit{et al.} propose DGG\cite{Yang_2019_ICCV} along similar lines with the additional constraint of minimizing the graph distances between embeddings of data points. 

Both ClusterGAN and VaDE require pre-training to initialize cluster centroids, and DGG requires pre-training to initialize the graph embeddings. Their success relies on the success of the initial pre-training. In this context, GVMAE learns the prior and posterior parameters jointly. However, their prior representation seems counter-intuitive as they use a sample from a normal distribution to generate the parameters (mean and variance) of the prior for each of the clusters. This implies that the prior for each class is dependent on a random variable rather than the class itself and hence would vary each time one would want to sample a latent for a class.

\section{Variational Autoencoders}
Variational Autoencoders (VAEs) \cite{rezende2014stochastic,kingma2013auto} are an application of Autoencoders for performing Variational Bayesian Inference over a set of data points. The main idea of Variational Inference is to learn a distribution in a latent space that can accurately capture the true distribution of the dataset. In particular, we wish to represent the joint probability $p(x, z)$ for points in dataset $x$ and their latent space representations $z$. This joint probability can be written as $p(x, z) = p(x|z)p(z)$ where $p(z)$ is a prior distribution from where latent variables are drawn and $p(x|z)$ is the conditional likelihood of a data point $x$ conditioned on the drawn latent variable $z$. The goal of variational inference is to infer the latent distribution from observed samples i.e. to accurately calculate $p(z|x)$. Using Bayes theorem, we can write this posterior distribution as:
\begin{equation}
    p(z|x) = \frac{p(x, z)}{p(x)} = \frac{p(x|z)p(z)}{p(x)}
\end{equation}

The problem is in approximating the evidence $p(x)$, inthe denominator, which requires an expensive marginalization over the latent variables. Therefore, Variational Inference seeks to approximate the posterior with a parameterized distribution $q_\theta(z|x)$ and minimize the KL-divergence between the true and approximated posterior w.r.t. to data distribution $p_{\mathcal{D}}(x)$.
{\small
\begin{equation}
\label{eq:post_kl}
    \mathbb{E}_{p_{\mathcal{D}}}KL(q_\theta(z|x)||p(z|x)) = \mathbb{E}_{q,p_{\mathcal{D}}}\log\frac{ q_\theta(z|x)}{p(x, z)} 
    + \mathbb{E}_{p_{\mathcal{D}}}\log p(x)
\end{equation}}

This can be re-written as 
{\small
\begin{equation}
\label{eq:evidence_log}
     \mathbb{E}_{p_{\mathcal{D}}}\log p(x) = \mathbb{E}_{p_{\mathcal{D}}}KL(q_\theta(z|x)||p(z|x)) + \mathbb{E}_{q,p_{\mathcal{D}}}\log\frac{p(x, z)}{q_\theta(z|x)}
\end{equation}}
By Jensen's inequality\cite{jensen1906fonctions}, the KL Divergence is always non-negative. Hence, the expectation in Eq. \ref{eq:evidence_log} acts as a lower bound for the evidence log-likelihood, and is hence called the Evidence Lower Bound (ELBO), given in Eq. \ref{eq:elbo}. 
\begin{equation}
\label{eq:elbo}
     ELBO = \mathbb{E}_{q,p_{\mathcal{D}}}\log \frac{p(x, z)}{q_\theta(z|x)}
\end{equation}
This allows us to write Eq. \ref{eq:evidence_log} as:
\begin{equation}
\label{eq:evidence_log2}
     \mathbb{E}_{p_{\mathcal{D}}}\log p(x) = \mathbb{E}_{p_{\mathcal{D}}}KL(q_\theta(z|x)||p(z|x)) + ELBO
\end{equation}
Therefore, minimizing the KL Divergence and hence, maximizing the log-likelihood of the evidence can be done by maximizing the ELBO. 
Since deep networks have strong representational abilities, Autoencoders have shown to perform well in approximating the distributions, leading to the birth of Variational Autoencoders (VAEs). Further information on Variational Inference and VAEs can be found in \cite{kingma2013auto}.

\begin{figure*}
    \centering
    \includegraphics[width=0.6\textwidth]{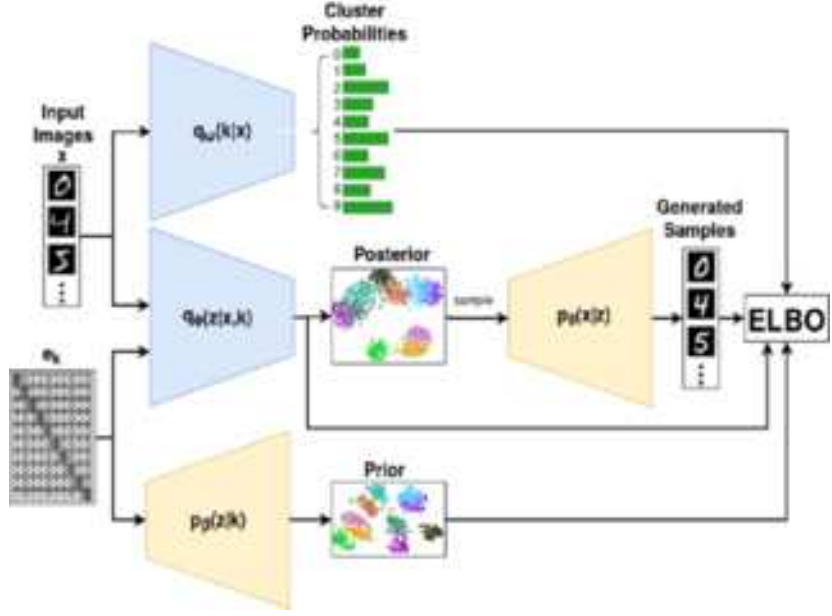}
    \caption{A flow chart of our training process. Given a set of input images $\mathbf{x}$ and its augments $\hat{\mathbf{x}}$, the posterior distribution $q_{\omega}(k|\mathbf{x})$ is first calculated, followed by $q_{\psi}(\mathbf{z}|\mathbf{x},k)$ and prior $p_{\beta}(\mathbf{z}|k)$, where $k$ is represented as $\mathbf{e}_k$, a one-hot vector having the  $k_{th}$ position is 1. We sample a latent from the posterior and generate a conditional prior $p_{\theta}(\mathbf{x}|\mathbf{z})$. We then calculate the ELBO according to Eq. \ref{eq:elbo_final}. Given the predictions for the images and its augments, we calculate the augmentation loss $\mathcal{L}_{aug}$ according to Eq. \ref{eq:aug_loss} and subsequently, calculate the final loss as given in Eq. \ref{eq:final_loss}. (Picture best viewed in colour)}
    \label{fig:flow}
\end{figure*}

\section{Proposed Approach}
VAEs traditionally model the prior as a single multivariate Gaussian. Since our objective is to cluster the data accurately, we use a Gaussian mixture prior for the latent space representation such that each Gaussian represents a cluster in the data. We aim to learn these Gaussians and an effective representation of the probability models, thereby learning the distribution a given data point belongs to, which corresponds to learning the clustering of the data. \par

An overview of our approach is shown in Fig. \ref{fig:flow}. Given a set of input images, we first predict their cluster probabilities. For each cluster, we infer a posterior latent distribution given the input and generate an independent prior distribution as well. We sample a latent from the posterior of each class from which we generate image samples which are used in computing the ELBO, along with the prior and the cluster probabilities.

\subsection{Generative Process}
\label{ssec:gen}
We first select a cluster $k$, from a categorical distribution characterized by $\pi$. we draw a sample $\mathbf{z}$ from the latent distribution $p(\mathbf{z}|k)$ of that cluster parameterized by a neural network $\beta$, to generate the conditional distribution $p(\mathbf{x}|\mathbf{z})$, which is parameterized by another neural network $\theta$. In mathematical terms, our generative model can be seen as 
\begin{gather}
\label{eq:prior}
p_{\beta,\theta}(\mathbf{x},\mathbf{z},k) = p_\theta(\mathbf{x}|\mathbf{z})p_\beta(\mathbf{z}|k)p(k)\\
p(k) = Cat(\pi)\\
p_\beta(\mathbf{z}|k) = \mathcal{N}(\mathbf{z}|\mathbf{\mu}_\beta(\mathbf{\mathbf{e}_k}), diag(\mathbf{\sigma}_\beta^2(\mathbf{\mathbf{e}_k}))) \\
p_\theta(\mathbf{x}|\mathbf{z}) = \mathcal{N}(\mathbf{x}|\mathbf{\mu}_\theta(\mathbf{z}), diag(\mathbf{\sigma}_\theta^2(\mathbf{z}))) \hspace{0.3em} or \hspace{0.3em} \mathcal{B}(\mathbf{x}|\mathbf{\mu}_\theta(\mathbf{z}))
\end{gather}
where $\pi = \frac{1}{K}$ is the categorical probability, $\mathbf{\mathbf{e}_k}$ is a one-hot vector with a 1 at the $k_{th}$ position, $\mathcal{N(.)}$ refers to a Normal distribution \& $\mathcal{B(.)}$ refers to a Bernoulli distribution. $\mathbf{\mu_\beta}, \mathbf{\sigma_\beta}^2, \mathbf{\mu_\theta}, \mathbf{\sigma_\theta}^2$ are the means \& variances, characterized by deep networks with parameters $\beta$ \& $\theta$ respectively.

Our process differs from GVMAE\cite{dilokthanakul2016deep} \& VaDE\cite{jiang2017variational} in the following ways. In GMVAE, a neural network samples a standard normal distribution as an input ($\mathbf{z}_1$ in Fig. \ref{fig:gmvae-gen}) \& generates a mean \& variance for each cluster ($\mathbf{z}_2$ in Fig. \ref{fig:gmvae-gen}). Therefore, the latent distribution for any given cluster depends on a normally distributed variable \& not just on the cluster it represents. Out of these, the $k^{th}$ mean \& variance is selected from which they sample their latent. 

In VaDE, the process is similar to ours however, they first train a stacked autoencoder with the input data \& then learn a GMM over the predicted latent representations from the encoder's bottleneck. This learnt GMM is used as the prior from which the latent variables are sampled. The performance of VaDE depends on how well the pre-training \& GMM initialization go, which would lead to a bad performance in case the initial learning doesn't properly converge. \par

\subsection{Inference Model}
\label{ssec:inf}
We perform variational inference by maximizing the Evidence Lower Bound ($ELBO$), which in turn leads to maximizing the log-likelihood of the evidence $\log p(x)$. With a slight abuse of notation, the ELBO can be written as:
\begin{equation}
\label{eq:gmvae_elbo}
    ELBO = \mathbb{E}_{q,p_{\mathcal{D}}}\log \frac{p_{\beta,\theta}(\mathbf{x},\mathbf{z},k)}{q_{\omega,\psi}(\mathbf{z},k|\mathbf{x})}
\end{equation}

where $\mathbb{E}_{q,p_{\mathcal{D}}}$ is the expectation with respect to $q_{\omega,\psi}(\mathbf{z},k|\mathbf{x})p_{\mathcal{D}}(\mathbf{x})$. Here, $q_{\omega,\psi}(\mathbf{z},k|\mathbf{x})$ is the approximate posterior of the inference model, which we factorize as:
\begin{equation}
\label{eq:post}
    q_{\omega,\psi}(\mathbf{z},k|\mathbf{x}) = q_{\omega}(k|\mathbf{x})q_{\psi}(\mathbf{z}|\mathbf{x},k)
\end{equation}
such that $\sum_{k=1}^{K} q_{\omega}(k|\mathbf{x}) = 1$ and  $q_{\psi}(\mathbf{z}|\mathbf{x},k) = \mathcal{N}(\mathbf{z}|\mathbf{\mu}_\psi(\mathbf{x},\mathbf{e}_k), diag(\mathbf{\sigma}_\psi^2(\mathbf{x},\mathbf{e}_k)))$
where $\mathbf{\mu_\psi}, \mathbf{\sigma_\psi}^2$ refer to the mean and variance of the posterior parameterized by a deep network $\psi$.

Our inference model is more straight-forward than GMVAE\cite{dilokthanakul2016deep} since they calculate their categorical posterior based on their latent's prior which, as mentioned towards the ending of Sec. \ref{ssec:gen}, depends on a sample from a standard normal distribution. This implies that the probability of a given category being assigned to a fixed category is stochastic in nature, which ideally shouldn't be the case. A latent variable is a representation of a sample data point. Therefore, the categorical posterior that depends on the latent, it should directly be inferred from the sampled data point itself. Our method directly predicts the categorical posterior $q_{\omega}(k|\mathbf{x})$ directly from the sampled data point $\mathbf{x}$. \par 

\subsection{Evidence Lower Bound}
\label{ssec:elbo}
Eq. \ref{eq:gmvae_elbo} can be factorized using Eq. \ref{eq:prior} \& \ref{eq:post} as:
\begin{equation}
ELBO = \mathbb{E}_{q,p_{\mathcal{D}}}[\log \frac{p_\theta(\mathbf{x}|\mathbf{z})p_\beta(\mathbf{z}|k)p(k)}{q_{\omega}(k|\mathbf{x})q_{\psi}(\mathbf{z}|\mathbf{x},k)}]
\label{eq:elbo_expanded}
\end{equation}

Since $\log p(k)$ is a constant, we drop it from Eq. \ref{eq:elbo_expanded} for ease of notation. Eq. \ref{eq:elbo_expanded} can then be re-written as:
\begin{equation}
\label{eq:elbo_final}
\begin{split}
ELBO = \mathbb{E}_{q,p_{\mathcal{D}}}[\log p_\theta(\mathbf{x}|\mathbf{z})]  - \mathbb{E}_{q_{\omega}(k|\mathbf{x})p_{\mathcal{D}}(\mathbf{x})} [\log q_{\omega}(k|\mathbf{x})] \\
\vspace{-10em}- \mathbb{E}_{q_{\omega}(k|\mathbf{x})p_{\mathcal{D}}(\mathbf{x})}[KL(q_{\psi}(\mathbf{z}|\mathbf{x},k)||p_\beta(\mathbf{z}|k))]
\end{split}
\end{equation}

In Eq. \ref{eq:elbo_final}, the first term $\mathbb{E}_{q,p_{\mathcal{D}}}[\log p_\theta(\mathbf{x}|\mathbf{z})]$ is the likelihood of a sampled data point $\mathbf{x}$ w.r.t. the generated distribution. It is similar to ensuring a proper reconstruction of the sampled input. It is further written as $\frac{1}{N}\sum_{i=1}^N\sum_{k=1}^K q_{\omega}(k|\mathbf{x}_i)\mathbb{E}_{\mathbf{z}\sim q_{\psi}(\mathbf{z}|\mathbf{x}_i,k)}[\log p_\theta(\mathbf{x}_i|\mathbf{z})]$, where $N$ is the batch size. This expansion allows us to circumvent the issue of having to sample from a discrete distribution which in turn would involve a discrete reparametrization. This expansion leaves us with having to sample a continuous latent $\mathbf{z}$ for each cluster $k$, which can easily be reparametrized for each $k$. The second term $\mathbb{E}_{q_{\omega}(k|\mathbf{x})p_{\mathcal{D}}(\mathbf{x})} [\log q_{\omega}(k|\mathbf{x})]$ is the expected entropy of the posterior distribution of $k$ w.r.t. the sampled data point $\mathbf{x}$, written as $\frac{-1}{N}\sum_{i=1}^N\sum_{k=1}^K q_{\omega}(k|\mathbf{x}_i)\log q_{\omega}(k|\mathbf{x}_i)$. 

The third and final term is the $\mathbf{z}$ KL term $\mathbb{E}_{q_{\omega}(k|\mathbf{x})p_{\mathcal{D}}(\mathbf{x})}[KL(q_{\psi}(\mathbf{z}|\mathbf{x},k)||p_\beta(\mathbf{z}|k))]$, which refers to the Kullback-Leibler Divergence of the prior w.r.t. the posterior distribution of $\mathbf{z}$. Since they are both Gaussians, the KL Divergence has a closed form solution.

One interesting point to note in the formulation is the anti-clustering nature of the $ELBO$, caused by the maximization of the entropy of $q_{\omega}(k|\mathbf{x})$. This regularization forces the information stored in $q_{\omega}(k|\mathbf{x})$ to be distributed among the clusters, rather than learning a one-hot prediction for the cluster labels. This is however mitigated by the reconstruction term, which would force the data to belong to a particular cluster. One advantage of this entropy maximization is that it prevents the trivial solution where all inputs are mapped to a single cluster, also known as "mode collapse" which is a common problem associated with VAEs. In our implementation, we found that the scale of the reconstruction term is exponentially larger than the KL term on real image data.
Hence, to ensure numerical stability, we multiply the reconstruction term in Eq. \ref{eq:elbo_final} with a scale-factor, $\lambda_{recons}$.

\subsection{Augmentation Loss}
\label{ssec:aug}
To make our model robust to input variations, we add augmented images along with the original images. The advantage of this is two-fold. Firstly, it allows the network to learn from a larger number of samples. Secondly, it allows us to use predictions of the original images to guide the predictions of the augmented images, resulting in a form of self-supervision. 
Different augments used are given in Sec. \ref{ssec:datasets}.

Let $\mathbf{x}$ be an input image and $\mathbf{\hat{x}}$, its augmented image. 
Intuitively, both should belong to the same cluster and posterior distribution. This enforces a strong constraint on the posterior predicted for $\mathbf{\hat{x}}$ by providing a form of supervision. Therefore, to ensure similarity in the predictions, we minimize the $L2$ distance between the predicted clusters and the expectation of the Wasserstein distance\cite{vaserstein1969markov} between the posteriors. 

\begin{equation}
\label{eq:aug_loss}
\begin{split}
    \mathcal{L}_{aug} = \sum_{k=1}^K||q_{\phi}(k|\mathbf{x}) - q_{\phi}(k|\mathbf{\hat{x}})||_2^2 \\
    + \mathbb{E}_{q_{\phi}(k|\mathbf{x})}[\mathcal{W}_2(q_{\psi}(\mathbf{z}|\mathbf{x},k), q_{\psi}(\mathbf{\hat{z}}|\mathbf{\hat{x}},k))]
\end{split}
\end{equation}

where $\mathcal{W}_2$ is the 2-Wasserstein distance. The expectation in the latter allows us to give a higher weight to the predicted cluster and a lower weight for clusters that are not associated with the given image. This has a closed form solution\cite{dowson1982frechet}, due to their Gaussian form, given by

\begin{equation}
\label{eq:w_loss}
\begin{split}
    \mathcal{W}_2(q_{\psi}(\mathbf{z}|x,k), q_{\psi}(\mathbf{\hat{z}}|\mathbf{\hat{x}},k)) = ||\mathbf{\mu}_\psi(\mathbf{x},\mathbf{e}_k) - \mathbf{\mu}_\psi(\mathbf{\hat{x}},\mathbf{e}_k)||_2^2\\
     + \hspace{0.2em} trace(C_1 + C_2 - 2(C_2^{\frac{1}{2}}C_1C_2^{\frac{1}{2}})^{\frac{1}{2}})
\end{split}
\end{equation}


the trace in Eq. \ref{eq:w_loss} can be simplified as:
\begin{equation}
\label{eq:trace_term}
\begin{split}
    C_1 + C_2 - 2(C_2^{\frac{1}{2}}C_1C_2^{\frac{1}{2}})^{\frac{1}{2}} &= diag(\mathbf{\sigma}_\psi^2(\mathbf{x},\mathbf{e}_k)) + \\ diag(\mathbf{\sigma}_\psi^2(\mathbf{\hat{x}},\mathbf{e}_k)) &- 2 diag(\mathbf{\sigma}_\psi(\mathbf{x},\mathbf{e}_k)\mathbf{\sigma}_\psi(\mathbf{\hat{x}},\mathbf{e}_k))\\
    = diag(\mathbf{\sigma}_\psi^2(\mathbf{x},\mathbf{e}_k)) + \mathbf{\sigma}_\psi^2(&\mathbf{\hat{x}},\mathbf{e}_k) - 2\mathbf{\sigma}_\psi(\mathbf{x},\mathbf{e}_k)\mathbf{\sigma}_\psi(\mathbf{\hat{x}},\mathbf{e}_k))\\
    = diag((\mathbf{\sigma}_\psi(\mathbf{x},\mathbf{e}_k) &- \mathbf{\sigma}_\psi(\mathbf{\hat{x}},\mathbf{e}_k))^2)\end{split}\end{equation}

Substituting Eq. \ref{eq:trace_term} in Eq. \ref{eq:w_loss}, we get
\begin{equation}
\label{eq:w_loss_mod}
\begin{split}
    \mathcal{W}_2(q_{\psi}(\mathbf{z}|\mathbf{x},k), q_{\psi}(\mathbf{\hat{z}}|\mathbf{\hat{x}},k)) =& \hspace{0.2em} ||\mathbf{\mu}_\psi(\mathbf{x},\mathbf{e}_k) - \mathbf{\mu}_\psi(\mathbf{\hat{x}},\mathbf{e}_k)||_2^2\\
     + \hspace{0.2em} trace(dia&g(\mathbf{\sigma}_\psi(\mathbf{x},\mathbf{e}_k) - \mathbf{\sigma}_\psi(\mathbf{\hat{x}},\mathbf{e}_k)^2))\\
    \
    \hspace{-3em}= ||\mathbf{\mu}_\psi(\mathbf{x},\mathbf{e}_k) - \mathbf{\mu}_\psi(\mathbf{\hat{x}},\mathbf{e}_k)||_2^2
    &+ ||\mathbf{\sigma}_\psi(\mathbf{x},\mathbf{e}_k) - \mathbf{\sigma}_\psi(\mathbf{\hat{x}},\mathbf{e}_k)||_2^2
\end{split}
\end{equation}

To prevent the augmentation from driving the initial learning phases, we anneal the effect of this loss so as to bring about its significance once a suitable amount of learning has progressed by modifying its weight $\lambda_{aug}$. Our final loss can be written as
\begin{equation}
\label{eq:final_loss}
    \mathcal{L} = -ELBO + \lambda_{aug}\mathcal{L}_{aug}
\end{equation}

\section{Experimental Details}
Details about the datasets used are given in Sec. \ref{ssec:datasets} and about the network architecture is given in the appendix in Sec. \ref{ssec:nets}. 
We implement the system using Tensorflow \cite{abadi2016tensorflow} using Adam \cite{kingma2014adam} as our optimizer with $beta1=0.9$, $beta2=0.999$ and the learning rate as $10^{-3}$ for MNIST \& Fashion-MNIST and $10^{-4}$ for the rest. The value of $\lambda_{aug}$ is 0.01 for the first 40 epochs, followed by 0.5 till 80 epochs and 1 thereafter. We use a batch-size of 100 for MNIST and Fashion-MNIST and 30 for the others. The value of the scale-factor for the reconstruction term, $\lambda_{recons}$, is kept at 0.5 for 40 epochs after which we keep it as 0.1 for CIFAR10 and STL-10. 

\begin{table*}[]
    \centering
    \noindent\resizebox{\textwidth}{!}{
    \begin{tabular}{|c|c||c|c|c|c|c|c|}
    \hline
        Method & Remarks & MNIST & Fashion-MNIST & STL-10 & CIFAR10 & CIFAR100 & FRGCv2\\
        \hline
        $k$-means on pixels & - & 53.49 & 47$^\dagger$ & 22.0 & 20.4 & - & -\\
        \hline
        IMSAT\cite{pmlr-v70-hu17b} & use ResNet\cite{he2016deep} & 98.4 & - & 94.1* & 45.6* & 27.5* & - \\ 
        \cdashline{2-2}
        VaDE\cite{jiang2017variational} & \multirow{3}{*}{Generative, use ResNet\cite{he2016deep}} & 94.46 & 57.8$^\ddagger$ & 84.5* & - & - & -\\ 
        Sarfaraz \textit{et al.}\cite{sarfraz2019efficient} & & 91.89 & - & 95.24* & - & - & -\\ 
        DDG\cite{Yang_2019_ICCV} & & 97.58 & - & 90.59* & - & - & -\\ 
        
        \arrayrulecolor{red}\hline\arrayrulecolor{black}
        Das \textit{et al.} \cite{das2019deep}& \multirow{3}{*}{Unsupervised, Pre-training} & \textbf{98.93} & - & - & 44.19 & 25.4 & 47.28\\ 
        Yang \textit{et al.} \cite{yang2019deep}& & 98 & 66.2 & - & - & - & -\\ 
        Nina \textit{et al.} \cite{Nina_2019_ICCV}& & 96.8 & 70.98 & - & 30.9 & - & -\\ 
        \cdashline{2-2}
        
        JULE\cite{yang2016joint} & \multirow{8}{*}{Purely Unsupervised}  & 96.1 & 56.3$^\ddagger$ & - & - & - & 46\\ 
        DEC\cite{xie2016unsupervised} & & 84.3 & 61.8$^\dagger$ & 35.9 & - & 14.3* & 37.8\\ 
        DEPICT\cite{dizaji2017deep} & & 96.3 & 39.2$^\ddagger$ & - & - & - & 47\\ 
        RDEC\cite{pmlr-v95-tao18a} & & 98.41 & - & 21.27 & - & - & - \\
        DEC-DA\cite{guo2018deep} & & 98.6 & 58.6 & - & - & - & -\\
        Haeusser \textit{et al.}\cite{haeusserassociative}& & 98.7 & - & 38.9 & 26.7 & - & 43.7\\ 
        \cline{2-2}
        GMVAE\cite{dilokthanakul2016deep} & \multirow{6}{*}{Generative, Purely Unsupervised} & 96.92 & 59.56$^\dagger$ & 25.36$^\dagger$ & 24.74$^\dagger$ & 10.10$^\dagger$ & 21.24$^\dagger$\\ 
        ClusterGAN\cite{mukherjee2019clustergan} & & 95 & 63 & - & - & - & -\\ 
        Ghasedi \textit{et al.}\cite{ghasedi2019balanced} & & 96.4 & - & 42.3 & 41.2 & - & 47.6\\ 
        
        \cline{1-1} \cline{3-8}
        \multirow{3}{*}{\textbf{Ours}} &  & 98.2 & \multirow{3}{*}{\textbf{71.72}} & \multirow{3}{*}{\textbf{43.9}} & \multirow{3}{*}{\textbf{44.5}} & \multirow{3}{*}{\shortstack{\textbf{25.6}\\($k=100$)}} & \multirow{3}{*}{\shortstack{\textbf{56.2}\\($k=20$)}}\\ 
        &  & 95.4 ($k=8)$ & & & & & \\ 
        &  & 98.4 ($k=20)$ & & & & & \\ 
        \hline
    \end{tabular}
}
    \caption{Clustering accuracy (\%). * - Use pre-trained features as input. $^\dagger$ - Results reported using source code. $^\ddagger$ - Results reported in \cite{guo2018deep}. We put a "-" where results are unavailable. For our results, we use $k=10$ except when specified. The methods above the red line, although unsupervised, use pre-trained features as input which itself boosts their performance be a big margin. Our main comparison is with unsupervised methods, shown below the red line. (Best viewed in colour)} 
    \label{tab:accuracy}
\vspace{-2em}
\end{table*}

\subsection{Datasets}
\label{ssec:datasets}
\begin{itemize}
    \item \textbf{Toy Data} used in the GMVAE paper, which is a 2D dataset consisting of 10000 points from the arcs of 5 circles. No augmentation is used for this. 
    
    \item \textbf{MNIST}\cite{lecun1998gradient} consists of 70000 black and white images of handwritten digits of size 28x28 split into 60000 training images and 10000 testing images. We use random rotations between 3$^\circ$ to -3$^\circ$ for the augmentation. 
    
    \item \textbf{Fashion MNIST}\cite{xiao2017fashion} consists of grayscale images of fashion items like sandals, T-shirts, handbag etc. It has the same size and split as MNIST but a greater complexity, making it more challenging. We add random brightness variations within a delta factor of 0.2 followed by an image binarization by checking if the pixel value is greater than a uniform random value.
    
    \item \textbf{STL-10}\cite{coates2011analysis} consists of 13000 labelled and 100000 unlabelled colour images of real world objects from ImageNet like cats, trucks, birds, ships etc. of size 96x96. The 13000 labelled images are split into 5000 training images and 8000 testing images. It is highly complex compared to MNIST and Fashion-MNIST. We augment the data using random shifts within a factor of 0.2 times the image dimensions, random rotations between 3$^\circ$ to -3$^\circ$, random shears within a factor of 0.2 and random vertical and horizontal flips.
    
    \item \textbf{CIFAR10} and \textbf{CIFAR100}\cite{krizhevsky2009learning} are diverse datasets of real world images. 9 classes in CIFAR10 are similar to STL-10. Achieving good clustering accuracy on them in an unsupervised manner is truly challenging. They consist of 50000 training and 10000 testing images each with 10 classes in CIFAR10 and 100 in CIFAR100. We use the same augmentations as in STL-10.
    
    \item \textbf{FRGCv2} consists of 2462 images having 20 face samples. It's a subset of the larger FRGC dataset. We use the subset provided by the authors of \cite{yang2016joint}, which is what other methods report their results on as well.
\end{itemize}
\subsection{Network Architechture}
\label{ssec:nets}
The network architectures for the prior and posterior networks are given in Tables \ref{tab:prior_dist} and \ref{tab:post_dist} respectively. The  $\mathbf{z}$ prior consists of fully connected networks as there aren't any images involved. The convolutional part of the posterior networks for CIFAR10, STL-10, CIFAR100 entail complex representations which are thereby, captured better by weight-sharing.

\section{Results}


\subsection{Clustering Results}
\begin{figure*}[h!]
	\centering
	\begin{tabular}{ccc}
	\centering
	\begin{subfigure}[b]{0.23\linewidth}
    	\centering
    	\frame{\includegraphics[width=\linewidth]{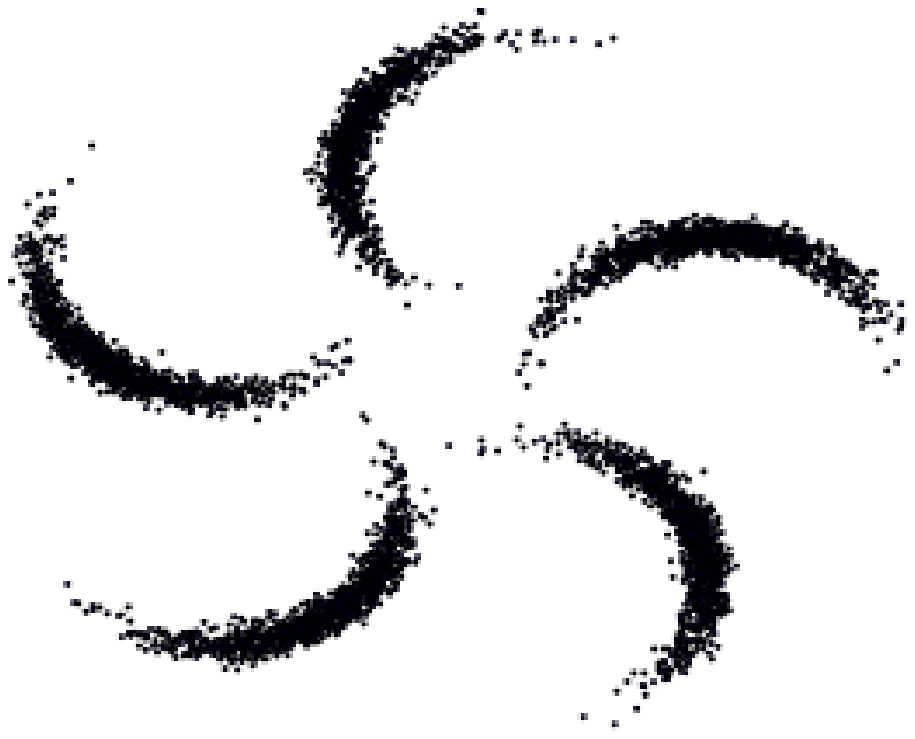}}
        \caption{Data Points}
    \end{subfigure}&
    \begin{subfigure}[b]{0.215\linewidth}
    	\centering
        \frame{\includegraphics[width=\linewidth]{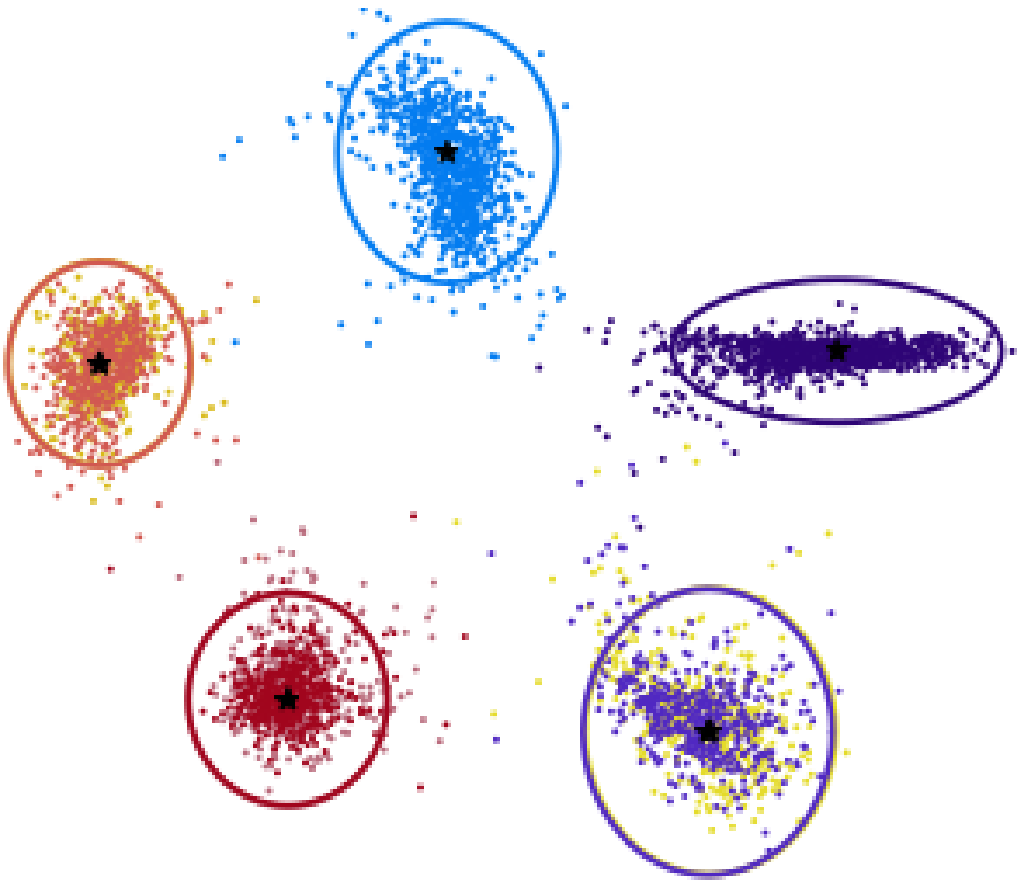}}
        \caption{GMVAE Latent space}
    \end{subfigure}& 
    \begin{subfigure}[b]{0.25\linewidth}
    	\centering
        \frame{\includegraphics[width=\linewidth]{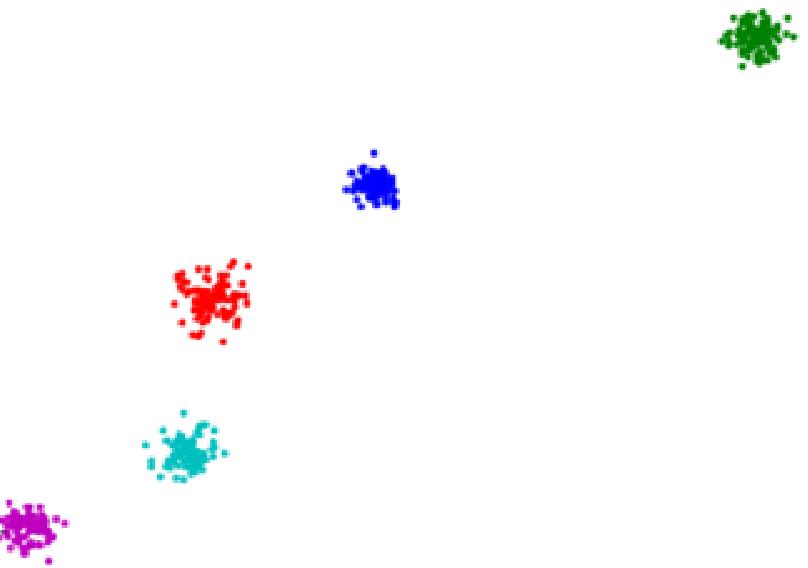}}
        \caption{Ours Latent space}
    \end{subfigure}
	\end{tabular}
	\caption{Toy dataset \& learnt latent spaces}
	\label{fig:toy_data}
\end{figure*}

\begin{table*}[]
\centering
\begin{subtable}[t]{\textwidth}
\centering
\resizebox{0.7\textwidth}{!}{
\begin{tabular}{|c|c|c|c|}
\hline
\textbf{Dataset} & Type of Layer & $p_\beta(\mathbf{z}|k)$ & $p_\theta(\mathbf{x}|\mathbf{z})$ \\ \hline
& Input & Input $\mathbf{e}_k$ {[}batch, K{]} & Input $\mathbf{z}$ {[}batch, 2{]} \\ \cline{2-4} 
\multirow{3}{*}{\begin{tabular}[c]{@{}c@{}}Toy Data\end{tabular}} & Hidden & FCN 120 & \begin{tabular}[c]{@{}c@{}}FCN 120\\ FCN 120\end{tabular} \\ \cline{2-4} 
& Output & \begin{tabular}[c]{@{}c@{}}$\mathbf{\mu}_\beta$, FCN 2, No activation\\ $\mathbf{\sigma}_\beta^2$, FCN 2, softplus activation\end{tabular} & \begin{tabular}[c]{@{}c@{}}$\mathbf{\mu}_\theta$, FCN 2, No activation\\ $\mathbf{\sigma}_\theta^2$, FCN 2, softplus activation\end{tabular} \\ \hline
 & Input & Input $\mathbf{e}_k$ {[}batch, K{]} & Input $\mathbf{z}$ {[}batch, 32{]} \\ \cline{2-4} 
\multirow{3}{*}{\begin{tabular}[c]{@{}c@{}}MNIST\\ Fashion Mnist\end{tabular}} & Hidden & -- & \begin{tabular}[c]{@{}c@{}} FCN 512 \\ FCN 2304, reshape {[}3,3,256{]}\\ UpConv 128 {[}3x3{]} \\ UpConv 128 {[}3x3{]}, stride=1, padding=valid \\UpConv 128 {[}3x3{]}, stride=1, padding=valid  \\UpConv 64 {[}3x3{]}, stride=1, padding=valid  \\UpConv 64 {[}3x3{]}\\UpConv 64 {[}3x3{]}, stride=1, padding=valid\end{tabular} \\ \cline{2-4} 
 & Output & \begin{tabular}[c]{@{}c@{}}$\mathbf{\mu}_\beta$, FCN 32, No activation\\ $\mathbf{\sigma}_\beta^2$, FCN 32, softplus activation\end{tabular} & $\mathbf{\mu}_\theta$, UpConv 1 {[}3x3{]}, stride=1, padding=valid, No activation \\ \hline
 & Input & Input $e_k$ {[}batch, K{]} & Input $\mathbf{z}$ {[}batch, 32{]} \\ \cline{2-4} 
 \multirow{3}{*}{\begin{tabular}[c]{@{}c@{}}CIFAR10\\ STL-10\\ CIFAR100\end{tabular}} & Hidden & \begin{tabular}[c]{@{}c@{}}FCN 64\\ FCN 64\end{tabular} & \begin{tabular}[c]{@{}c@{}}FCN  3072, reshape {[}4,4,192{]}\\ UpConv 192 {[}3x3{]}, padding=valid\\ UpConv 192 {[}3x3{]}, stride=1, padding=valid\\ UpConv 96 {[}3x3{]}, stride=1, padding=valid \\ UpConv 96 {[}3x3{]}, padding=valid,\\ UpConv 96 {[}3x3{]}, stride=1, padding=valid \\ UpConv 96 {[}3x3{]} stride=1, padding=valid \end{tabular} \\ \cline{2-4} 
 & Output & \begin{tabular}[c]{@{}c@{}}$\mathbf{\mu}_\beta$, FCN 32, No activation\\ $\mathbf{\sigma}_\beta^2$, FCN 32, softplus activation\end{tabular} & \begin{tabular}[c]{@{}c@{}}$\mathbf{\mu}_\theta$, UpConv 3 {[}3x3{]}, stride=1, padding=valid, No activation\\ $\mathbf{\sigma}_\theta^2$, UpConv 3 {[}3x3{]}, stride=1, padding=valid, softplus activation\end{tabular} \\ \hline
 & Input & Input $e_k$ {[}batch, K{]} & Input $\mathbf{z}$ {[}batch, 32{]} \\ \cline{2-4} 
 \multirow{3}{*}{\begin{tabular}[c]{@{}c@{}}FRGCv2\end{tabular}} & Hidden & \begin{tabular}[c]{@{}c@{}}FCN 64\end{tabular} & \begin{tabular}[c]{@{}c@{}}FCN 512, reshape {[}4,4,32{]}\\ UpConv 64 {[}3x3{]}\\ UpConv 32 {[}3x3{]}\end{tabular} \\ \cline{2-4} 
 & Output & \begin{tabular}[c]{@{}c@{}}$\mathbf{\mu}_\beta$, FCN 32, No activation\\ $\mathbf{\sigma}_\beta^2$, FCN 32, softplus activation\end{tabular} & \begin{tabular}[c]{@{}c@{}}$\mathbf{\mu}_\theta$, UpConv 1 {[}5x5{]}, No activation\\ $\mathbf{\sigma}_\theta^2$, UpConv 1 {[}5x5{]}, softplus activation\end{tabular} \\ \hline
\end{tabular}
}
\subcaption[]{Network architecture for the prior probabilities $p_\beta(\mathbf{z}|k)$ \& $p_\theta(\mathbf{x}|\mathbf{z})$.} 
    \label{tab:prior_dist}
\end{subtable}

\begin{subtable}[t]{\textwidth}
\centering
\resizebox{0.6\textwidth}{!}{
\begin{tabular}{|c|c|c|c|}
\hline
\textbf{Dataset} & Type of Layer & $q_{\omega}(k|\mathbf{x})$ & $q_{\psi}(\mathbf{z}|\mathbf{x},k)$ \\ \hline
& Input & Input $\mathbf{x}$ {[}batch, 2{]} & Input $\mathbf{x}$ {[}batch, 2{]}, $\mathbf{e}_k$ {[}batch, K{]}\\ \cline{2-4} 
\multirow{3}{*}{\begin{tabular}[c]{@{}c@{}}Toy Data\end{tabular}} & Hidden & \begin{tabular}[c]{@{}c@{}}FCN 120\\    FCN $K$ No activation\end{tabular} & \begin{tabular}[c]{@{}c@{}}FCN 10\\ concat $\mathbf{e}_k$\\    FCN 10\end{tabular} \\ \cline{2-4} 
 & Output & $\mathbf{\hat{e}}_k$, softmax activation & \begin{tabular}[c]{@{}c@{}}$\mathbf{\mu}_\psi$, FCN 2, No activation\\ $\mathbf{\sigma}_\psi^2$, FCN 2, softplus activation\end{tabular} \\ \hline
 & Input & Input $\mathbf{x}$ {[}batch, 28, 28, 1{]} & Input $\mathbf{x}$ {[}batch, 28, 28, 1{]}, $\mathbf{e}_k$ {[}batch, K{]}\\ \cline{2-4} 
\multirow{4}{*}{\begin{tabular}[c]{@{}c@{}}MNIST\\ Fashion Mnist\end{tabular}}  & \multirow{2}{*}{Hidden} & \multicolumn{2}{c|}{\begin{tabular}[c]{@{}c@{}}Conv 64 {[}3x3{]}, stride=1, padding=valid\\ Conv 64 {[}3x3{]}, stride=1, padding=valid\\ Conv 64 {[}3x3{]}, stride=1, padding=valid\\ Conv 128 {[}3x3{]}, padding=valid \\ Conv 128 {[}3x3{]}, stride=1, padding=valid \\ Conv 128 {[}3x3{]}, stride=1, padding=valid\\ Conv 256 {[}3x3{]}, padding=valid \end{tabular}}\\ \cline{3-4} 
& & \begin{tabular}[c]{@{}c@{}}Flatten\\ FCN 512, Dropout 0.8\\ FCN 512, Dropout 0.6\\ FCN K, No activation\end{tabular} & \begin{tabular}[c]{@{}c@{}}Flatten, concat $\mathbf{e}_k$\\ FCN 256\\ FCN 512\end{tabular} \\ \cline{2-4} 
 & Output & $\mathbf{\hat{e}}_k$, softmax activation & \begin{tabular}[c]{@{}c@{}}$\mathbf{\mu}_\psi$, FCN 32, No activation\\ $\mathbf{\sigma}_\psi^2$, FCN 32, softplus activation\end{tabular} \\ \hline
 & Input & Input $\mathbf{x}$ {[}batch, 32, 32, 3{]} & Input $\mathbf{x}$ {[}batch, 32, 32, 3{]}, $\mathbf{e}_k$ {[}batch, K{]}\\ \cline{2-4} 
 \multirow{4}{*}{\begin{tabular}[c]{@{}c@{}}CIFAR10\\ STL-10\\ CIFAR100\end{tabular}} & \multirow{2}{*}{Hidden} & \multicolumn{2}{c|}{\begin{tabular}[c]{@{}c@{}}Conv 96 {[}3x3{]}, stride=1, padding=valid\\ Conv 96 {[}3x3{]}, stride=1, padding=valid\\ Conv 96 {[}3x3{]}, padding=valid\\ Conv 192 {[}3x3{]}, stride=1, padding=valid \\ Conv 192 {[}3x3{]}, stride=1, padding=valid \\ Conv 192 {[}3x3{]}, padding=valid\end{tabular}}\\ \cline{3-4} 
  & & \begin{tabular}[c]{@{}c@{}}Flatten\\ FCN 128, Dropout 0.8\\ FCN 64, Dropout 0.99\\ FCN K, No activation\end{tabular} & \begin{tabular}[c]{@{}c@{}}Flatten, concat $\mathbf{e}_k$\\ FCN 512\\ FCN 512\end{tabular} \\ \cline{2-4} 
 & Output & $\mathbf{\hat{e}}_k$, softmax activation & \begin{tabular}[c]{@{}c@{}}$\mathbf{\mu}_\psi$, FCN 32, No activation\\ $\mathbf{\sigma}_\psi^2$, FCN 32, softplus activation\end{tabular} \\ \hline
 & Input & Input $\mathbf{x}$ {[}batch, 32, 32, 3{]} & Input $\mathbf{x}$ {[}batch, 32, 32, 3{]}, $\mathbf{e}_k$ {[}batch, K{]}\\ \cline{2-4} 
 \multirow{4}{*}{\begin{tabular}[c]{@{}c@{}}FRGCv2\end{tabular}} & \multirow{2}{*}{Hidden} & \multicolumn{2}{c|}{\begin{tabular}[c]{@{}c@{}}Conv 32 {[}5x5{]}\\ Conv 64 {[}3x3{]}\\ Conv 128 {[}3x3{]}\\ Conv 32 {[}1x1{]}\end{tabular}}\\ \cline{3-4} 
  & & \begin{tabular}[c]{@{}c@{}}Flatten\\ FCN 64, Dropout 0.9\\ FCN 64, Dropout 0.7\\ FCN K, No activation\end{tabular} & \begin{tabular}[c]{@{}c@{}}Flatten, concat $\mathbf{e}_k$\\ FCN 512\\ FCN 512\end{tabular} \\ \cline{2-4} 
 & Output & $\mathbf{\hat{e}}_k$, softmax activation & \begin{tabular}[c]{@{}c@{}}$\mathbf{\mu}_\psi$, FCN 32, No activation\\ $\mathbf{\sigma}_\psi^2$, FCN 32, softplus activation\end{tabular} \\ \hline
\end{tabular}
}
\subcaption[]{Network architecture for the prior probabilities $p_\beta(\mathbf{z}|k)$ \& $p_\theta(\mathbf{x}|\mathbf{z})$.} 
    \label{tab:post_dist}
\end{subtable}

\caption{Network architectures for (a) posterior probabilites and (b) prior probabilities. All the Conv and UpConv layers have a stride of 2 with padding as 'same' unless specified otherwise. For MNIST and Fashion MNIST, all layers have ReLU activations unless specified. For others, all layers have leaky ReLU activations unless specified. $K$ - number of clusters (Toy Data - 5, MNIST, Fashion MNIST, STL-10, CIFAR10 - 10, FRGCv2 - 20, CIFAR100 - 100)}
\label{tab:nets}
\end{table*}
Firstly, to see the representation capability of our method, we show the latent space learnt by our method \& GMVAE on the synthetic dataset given in their paper. As seen in Fig. \ref{fig:toy_data}, our clusters are tightly packed individually \& more spread overall compared to GMVAE. Good latent space separation is key for clustering as it can partition the data in a better way.

\begin{figure*}
\centering
        \begin{subfigure}[b]{0.2\textwidth}
                \includegraphics[width=\linewidth]{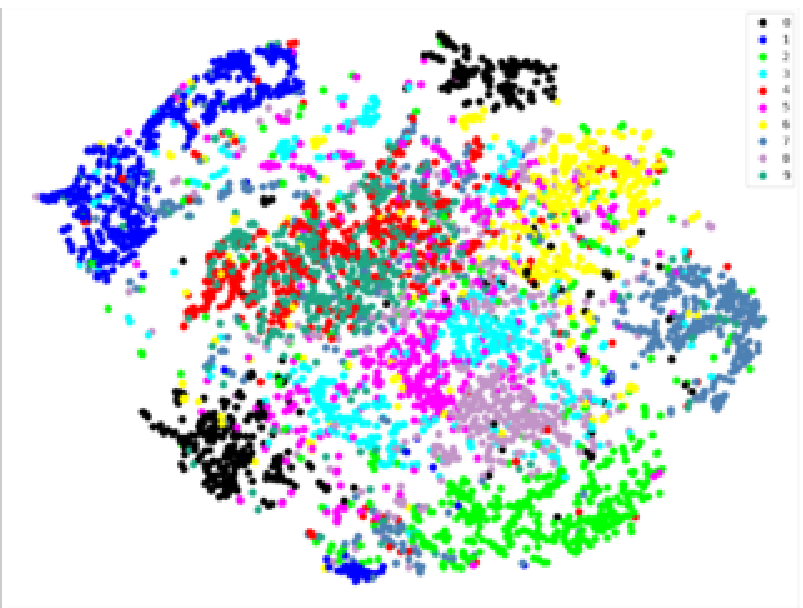}
                \caption{Epoch 0}
        \end{subfigure}%
        \begin{subfigure}[b]{0.2\textwidth}
                \includegraphics[width=\linewidth]{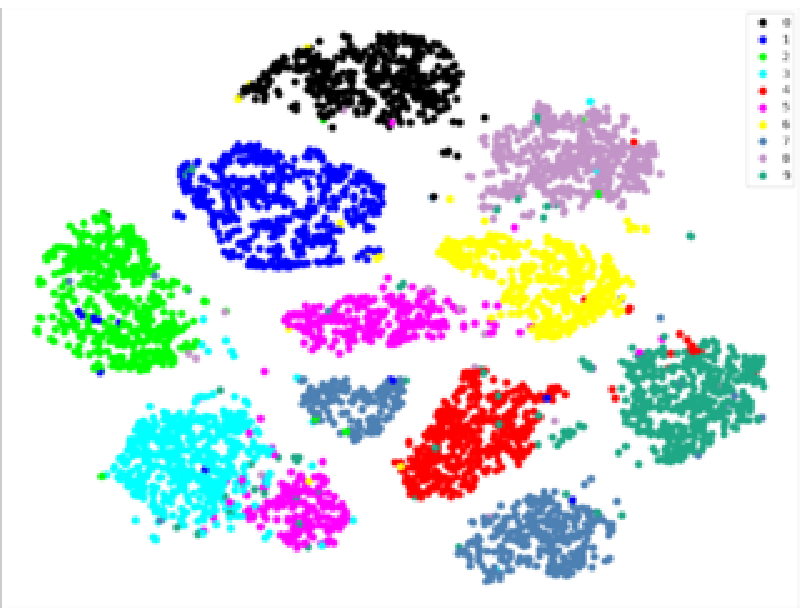}
                \caption{Epoch 22}
        \end{subfigure}%
        \begin{subfigure}[b]{0.2\textwidth}
                \includegraphics[width=\linewidth]{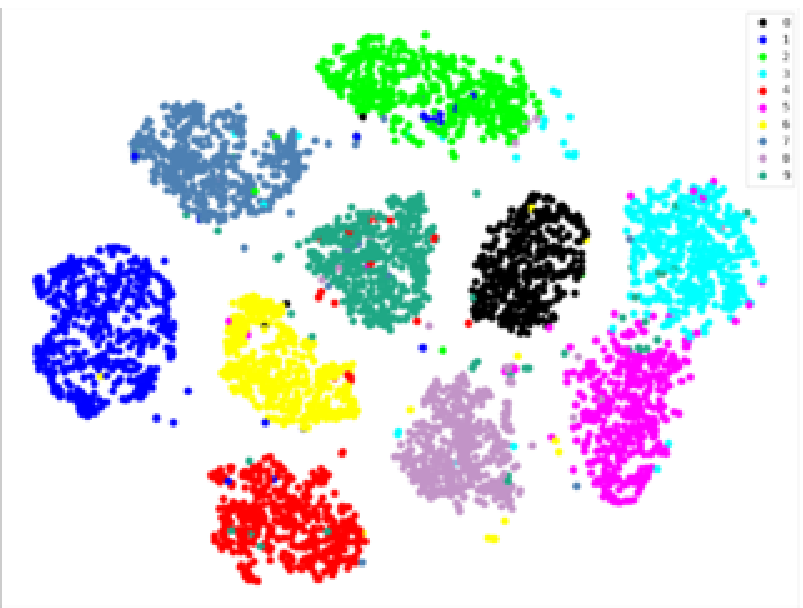}
                \caption{Epoch 44}
        \end{subfigure}%
        \begin{subfigure}[b]{0.2\textwidth}
                \includegraphics[width=\linewidth]{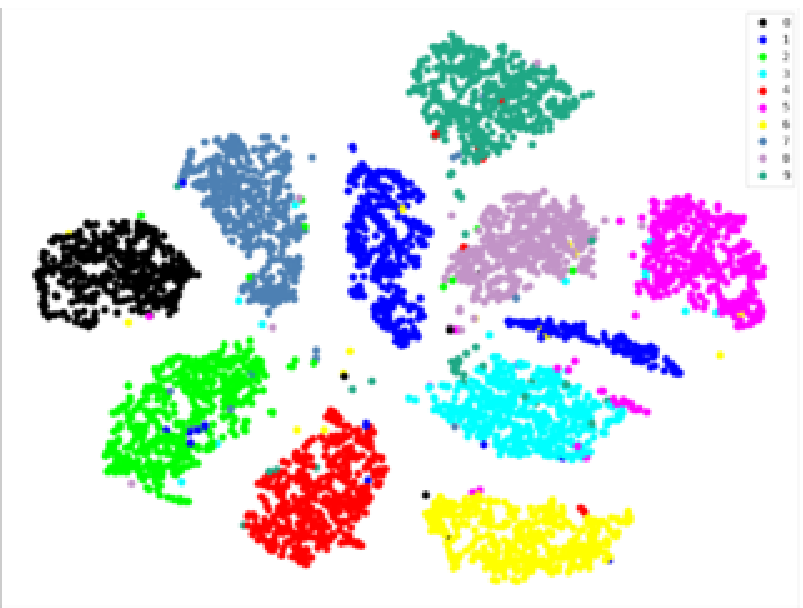}
                \caption{Epoch 66}
        \end{subfigure}%
        
        \caption{t-SNE Visualization of posterior latent representation of MNIST samples during training}\label{fig:tsne_post}
\end{figure*}

\begin{figure*}[h!]
\centering
        \begin{subfigure}[b]{0.2\textwidth}
                \includegraphics[width=\linewidth]{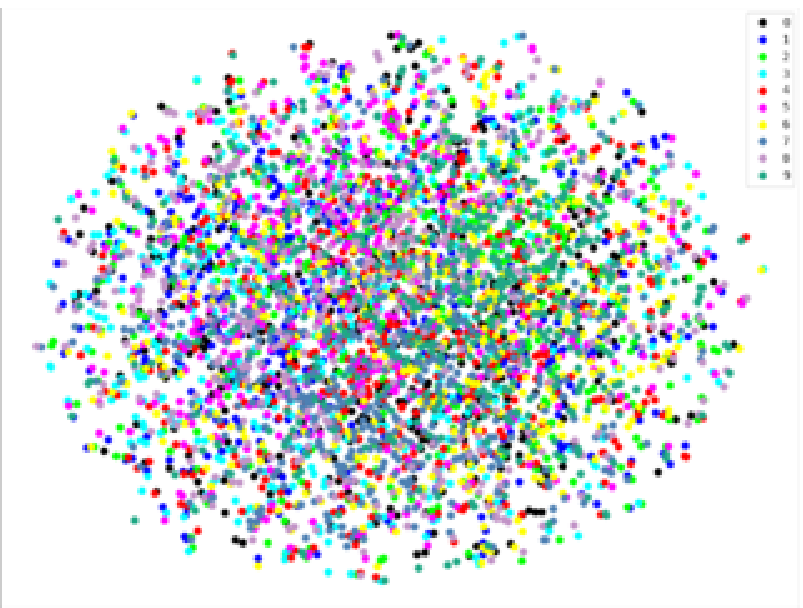}
                \caption{Epoch 0}
        \end{subfigure}%
        \begin{subfigure}[b]{0.2\textwidth}
                \includegraphics[width=\linewidth]{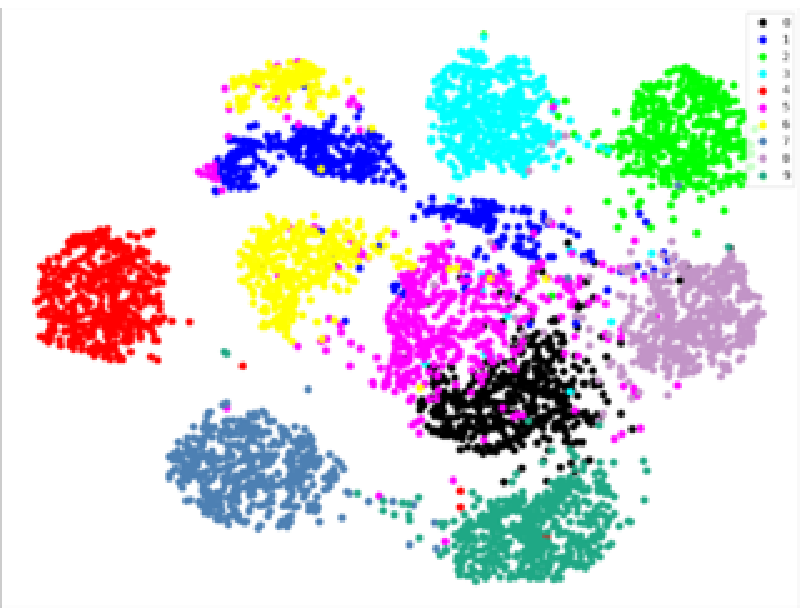}
                \caption{Epoch 22}
        \end{subfigure}%
        \begin{subfigure}[b]{0.2\textwidth}
                \includegraphics[width=\linewidth]{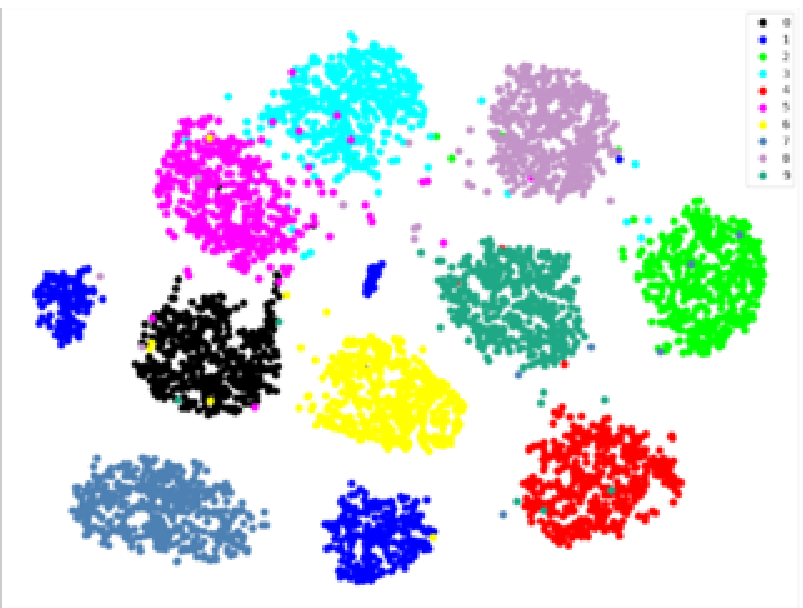}
                \caption{Epoch 44}
        \end{subfigure}%
        \begin{subfigure}[b]{0.2\textwidth}
                \includegraphics[width=\linewidth]{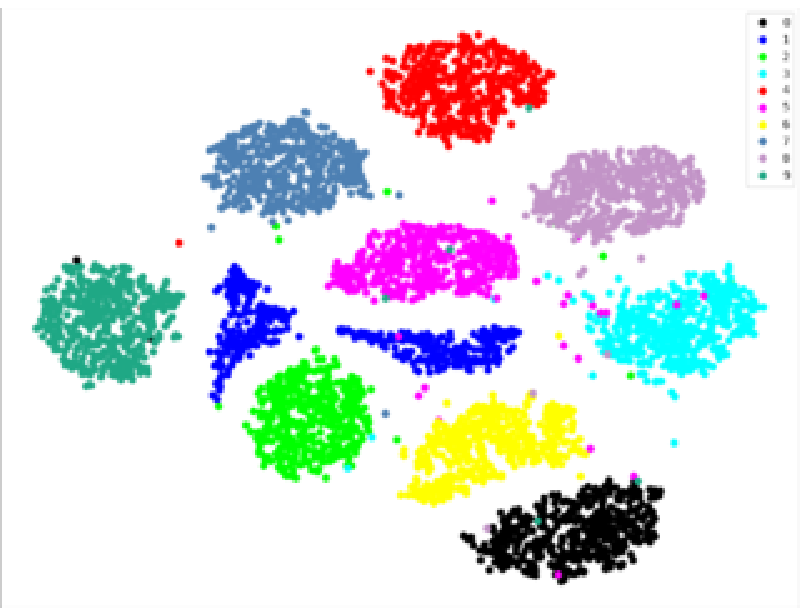}
                \caption{Epoch 66}
        \end{subfigure}%
        \caption{t-SNE Visualization of samples drawn from the learnt prior latent distribution of MNIST during training}\label{fig:tsne_prior}
\end{figure*}

Furthermore, to see how well the clusters are separated in the latent space, we visualize the means of the posterior and the prior using t-SNE\cite{maaten2008visualizing}. The posterior representation of a subset of testing images can be seen in Fig. \ref{fig:tsne_post}. For the prior, we randomly generate 500 samples from each cluster, to see how well the prior has been learnt. This can be seen in Fig. \ref{fig:tsne_prior}. Evidently, we can see a clear separation between clusters from the beginning itself (11 epochs), which gets refined as the iterations go on. This not only shows the discriminatory nature of the learnt representations, but also the speed with which it becomes a characteristic of the latent space. 



For evaluating our method, we use the \textit{unsupervised clustering accuracy (ACC)}\cite{xie2016unsupervised}, which can be given as
\begin{equation}
    ACC = \max_{m\in\mathcal{M}}\frac{\sum_{i=1}^N 1(l_i == m(c_i))}{N}
\end{equation}
where $N$ is the total no. of images, $l_i$ and $c_i$ are the label and cluster for the $i^{th}$ image respectively and $\mathcal{M}$ is the set of possible one-to-one mappings between clusters and labels.\par

Our accuracy can be seen in Table \ref{tab:accuracy}. It is important to note that we only lag behind methods which use pre-trained ResNet\cite{he2016deep} features as input, which itself is an extremely informative representation, giving them a large advantage as it has already covered thousands of samples of the classes present in STL-10 and CIFAR10 in a supervised manner. In comparison with Das \textit{et al.}\cite{das2019deep} who use a lot of epochs for pre-training, we perform better especially on more complex datasets like Fashion-MNIST, STL-10, CIFAR10 and CIFAR100.\par

Our main comparison is with unsupervised methods, shown below the red line in Table \ref{tab:accuracy}. Our performance significantly exceeds them in all the complex real world datasets achieving state-of-the-art results among unsupervised methods. Our performance exceeds that of generative methods\cite{mukherjee2019clustergan,ghasedi2019balanced} on all datasets, showing that the inference power of VAEs is greater than GANs for clustering, where a strong representation of the data distribution is important. We perform better than GMVAE\cite{dilokthanakul2016deep} and VaDE\cite{jiang2017variational} on all datasets showing the effectiveness of our approach to using a Gaussian Mixture latent space.\par 


\subsection{STL-CIFAR Transfer Learning}
\label{ssec:transfer_leanring}
CIFAR10 and STL-10 are inherently similar, having 9 same classes out of 10 and has some scope for transfer learning. To show such a capability of our method, we test the model trained on CIFAR10 using STL-10 and vice-versa on these common classes. The results of this are shown in Table \ref{tab:transfer_leanring}.
\begin{table}[h!]
	\centering
	\begin{tabular}{|c|c||c|}
	\hline
	\textbf{Trained on} & 	\textbf{Tested on} & \textbf{Accuracy}\\
	\hline

	STL-10 & CIFAR10 & \textbf{38.8}\\ 
	\hline
	CIFAR10 & STL-10 & \textbf{41.03}\\
	\hline
	\end{tabular}
    \caption{Transfer learning between CIFAR10 and STL-10}
	\label{tab:transfer_leanring}
		
\end{table}

\begin{table}
    \centering
\noindent\resizebox{0.45\textwidth}{!}{
    \begin{tabular}{|c|c||c|c|c|c|}
    \hline
        \diagbox{Method}{$r_{min}$} & 0.1 & 0.3 & 0.5 & 0.7 & 0.9\\
        \hline
    $k$-means & 47.14 & 49.93 & 53.65 & 54.16 & 54.39\\
        AE+$k$-means & 66.82 & 74.91 & 77.93 & 80.04 & 81.31\\ 
        DEC & 70.10 & 80.92 & 82.68 & 84.69 & 85.41\\ 
        \hline
        \textbf{Ours} & \textbf{89.6} & \textbf{90.3} & \textbf{96.1} & \textbf{96.9} & \textbf{97.1}\\
        \hline
    \end{tabular}}
    \caption{Clustering accuracy (\%) for MNIST with imbalanced clusters using different min. retention rates $r_{min}$.} 
    \label{tab:imbalanced}
    \vspace{-2em}
\end{table}
\subsection{Imbalanced Clusters}
To study our performance on imbalanced data, we follow the imbalanced sampling mechanism as in DEC\cite{xie2016unsupervised} for MNIST. We define  which is 
the probability of retaining samples of class 0 is defined as a minimum retention rate $r_{min}$, for class 9 the retention probability is 1 and the for other classes, it varies linearly from $r_{min}$ to 1. The ratio of the sizes of the largest and smallest clusters is $1/r_{min}$. We perform significantly better than other methods as shown in Table \ref{tab:imbalanced}.


\subsection{Generated Samples}
In order to see the effectiveness of our latent space representation, we generate images to see if captures the underlying data distribution in a meaningful manner. Since our approach is inherently a generative one, it should be able to extrapolate samples from the latent space, as is the case with any generative approaches. The ability to generate high quality samples is the true test for any generative approach.\par

\begin{figure}[h!]
    \centering
    \begin{subfigure}[b]{0.2\textwidth}
                \includegraphics[width=0.95\linewidth]{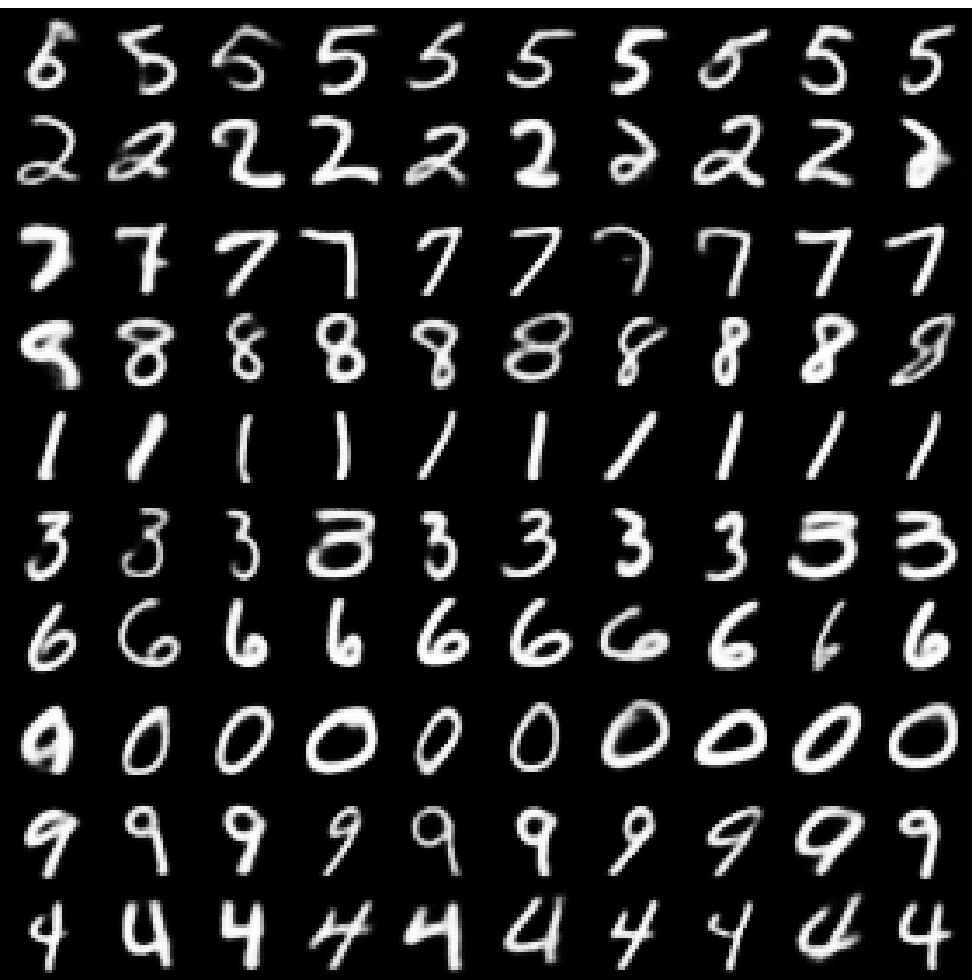}
                \caption{MNIST}
                \label{fig:gen_mnist}
        \end{subfigure}%
    \begin{subfigure}[b]{0.2\textwidth}
                \includegraphics[width=0.95\linewidth]{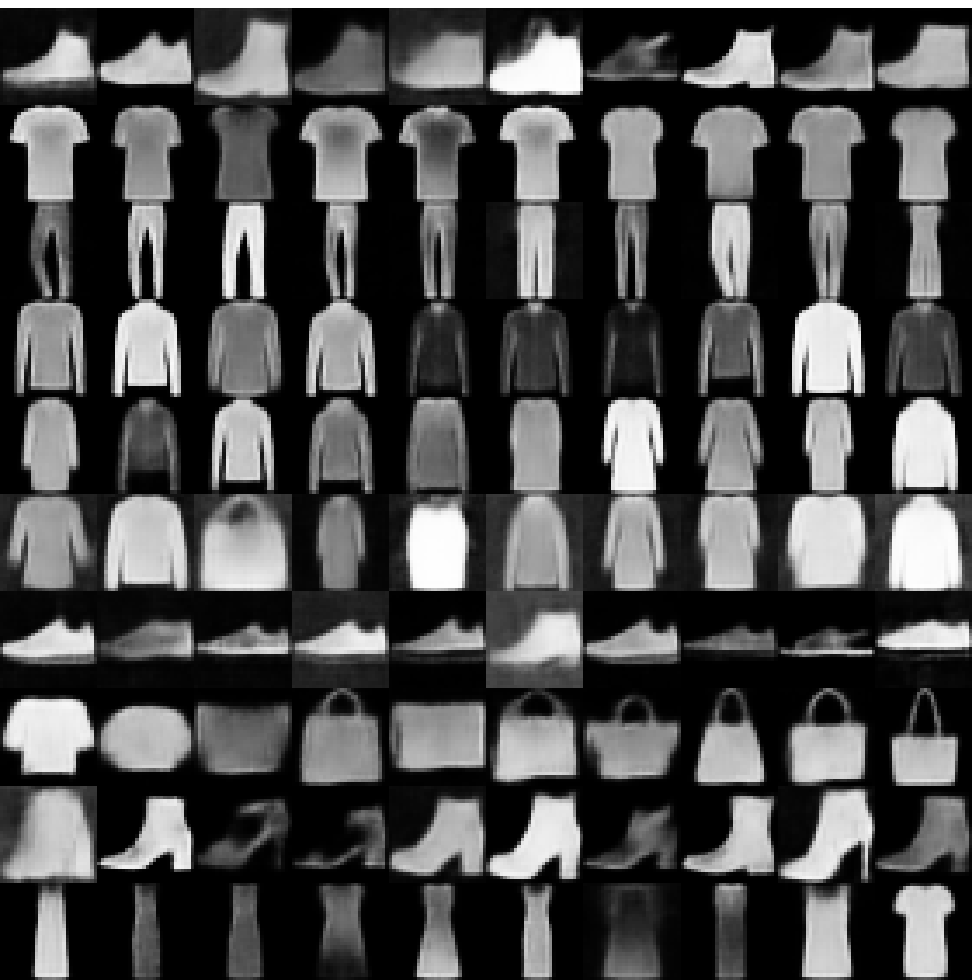}
                \caption{Fashion MNIST}
                \label{fig:gen_fashion}
        \end{subfigure}%
  
    \caption{Images generated from each learned cluster. Each row represents a cluster and each column is an independent Bernoulli distribution $p_\theta(\mathbf{x}|\mathbf{z})$ given a $\mathbf{z}$ sampled from the Gaussian Distribution $p_\beta(\mathbf{z}|k)$ of the $k^{th}$ cluster.}
    \label{fig:gen}
\vspace{-1em}    
\end{figure}

As it can be seen in Fig. \ref{fig:gen}, our learned model is able to generate realistic samples from each cluster. Moreover, one more important inference that can be drawn is that our method learns to distinguish properly between classes since it is able to generate distinct samples from each of the clusters. This shows that our generative approach works well not only in distinguishing between different clusters but also in being able to generate realistic samples given latent data.

\subsection{Latent Space Interpolation}
To see how well the learning has progressed, we visualize how variations in the latent space correspond to variations in generated samples. We perform 2 types of interpolations, one in the continuous $\mathbf{z}$ space and one in the $k$ space. Though $k$ is a discrete variable, we vary it continuously to see how data generation for unknown inputs looks like. 
In our model, we have two sets of variables on which the output of the generative model depends on. The first is the categorical distribution for $k$, followed by a Gaussian distribution corresponding to the $k^{th}$ cluster, for obtaining $\mathbf{z}$. We generate samples by interpolating in both these spaces to see how the generated samples are, giving us further insight into the extent to which each of these variables influence the generation process.
\subsubsection{$\mathbf{z}$-Interpolation}
We choose two clusters $k_1$ \& $k_2$ and sample latents $\mathbf{z}_1$ \& $\mathbf{z}_2$ from their respective Gaussians. We calculate our interpolated latent as $\mathbf{\Tilde{z}} = \alpha \mathbf{z}_1 + (1-\alpha)\mathbf{z}_2$ by varying $\alpha$ from 0 to 1, from which we generate our output.

As shown in Fig. \ref{fig:interp_z}, smooth variations in the latent space generate smooth transitions in the image space. This shows that our distributions in the latent space capture the data distribution accurately. This is an important result showing the ability of our method to accurately capture changes in the latent space and reflect them in the input space. 

\subsubsection{$k$-Interpolation}
We choose two clusters $k_1$ and $k_2$ and linearly interpolate between their one-hot vector representations $\mathbf{e_{k_1}}$ and $\mathbf{e_{k_2}}$. We generate $\mathbf{\Tilde{e}} = \alpha \mathbf{e_{k_1}} + (1-\alpha)\mathbf{e_{k_2}}$ by varying $\alpha$ from 0 to 1. From $\mathbf{\Tilde{e}}$, we generate the parameters of the prior for sampling $\mathbf{z}$ from which we generate our output. 
\begin{figure}
    \centering
    \begin{subfigure}[b]{0.2\textwidth}
        \includegraphics[width=0.95\linewidth]{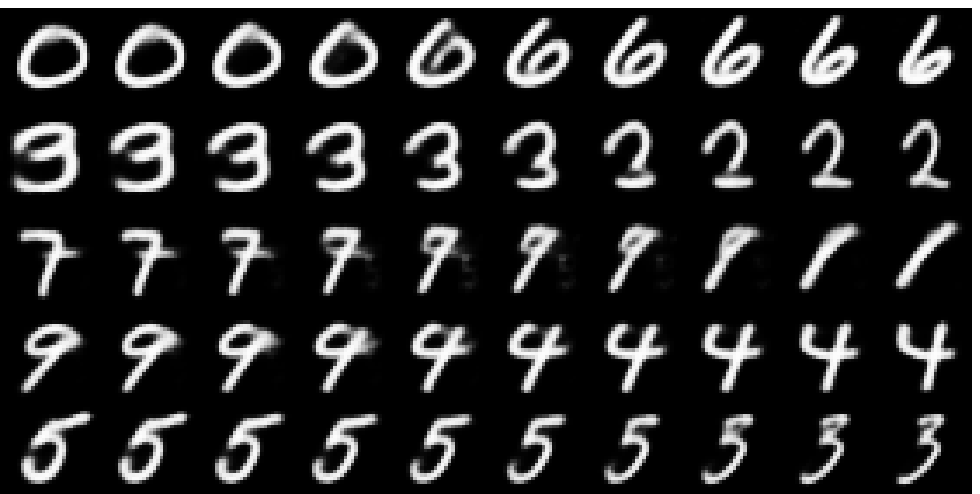}
        \caption{$\mathbf{z}$-Interpolation}
        \label{fig:interp_z}
        \end{subfigure}%
    \begin{subfigure}[b]{0.2\textwidth}
        \includegraphics[width=0.95\linewidth]{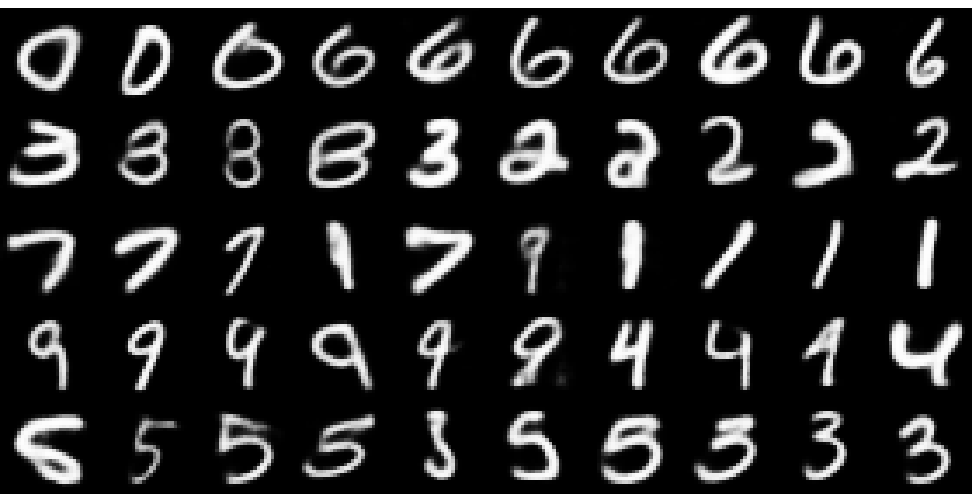}
        \caption{$k$-Interpolation}
        \label{fig:interp_k}
        \end{subfigure}%
    

    \caption{Images generated by interpolating $\mathbf{z}$ and $k$. The first column is for $\alpha = 0$ and the last for $\alpha = 1$.}
    \label{fig:interp}
    
\end{figure}

From Fig. \ref{fig:interp_k}, it can be said that our method is able to extrapolate data not just from continuous variations in the latent space but more importantly from continuous variations in a discrete space. The main takeaway is that though we train our model using only one-hot vectors, our method is able generate meaningful images representing the the interpolated vector $\mathbf{\Tilde{e}}$. This result shows that our model is able to learn a truly effective representation for each cluster, where the data distribution is preserved not just in the latent space ($\mathbf{z}$) but in the categorical space ($k$) as well. The $k$-interpolation is not as smooth as the $\mathbf{z}$-interpolation mainly because the former is a discrete space whereas the latter is a smooth continuous one.

\begin{table}[]
    \centering
    \resizebox{.4\textwidth}{!}{
    \begin{tabular}{|c|c|c|c|}
    \hline
        Method  & MNIST & Fashion-MNIST & CIFAR-10\\
        \hline
        \textbf{VAE with GMM prior} & 96.8 & 65.3 & 37.1\\ 
        \textbf{With $\lambda_{recons}$ = 0.4} & - & - & 38.5\\
        \textbf{With $\lambda_{recons}$ = 0.5} & - & - & 39.1\\
        \textbf{With $\lambda_{recons}$ = 0.65} & - & - & 41.8\\
        \textbf{With $\lambda_{recons}$ = 0.7} & - & - & 39.2\\
        \textbf{With $L2$ augmentation} & 97.2 & 66.4 & 42.7\\ 
        \textbf{With $\mathcal{W}_2$ augmentation} & 97.7 & 67.6 & 43.8\\ 
        \hline
        \textbf{Ours-final $k=8$} & 95.4 & - & -\\ 
        \textbf{Ours-final $k=10$} & 98.2 & \textbf{71.72} & \textbf{44.5}\\ 
        \textbf{Ours-final $k=20$} & \textbf{98.4} & - & -\\ 
        \hline
    \end{tabular}
    }
    \caption{Ablation study of our proposed method done by incorporating each of our proposed additions individually. We report the clustering accuracy (\%) of each variant.} 
    \label{tab:ablation}
    \vspace{-2em}    
\end{table}
\subsection{Ablation study}
We perform an ablation study to understand the effect of different parts of our loss where we compare the following:
\begin{itemize}
    \item A vanilla implementation of our proposed VAE.
    \item Adding a scale factor to the reconstruction term.
    \item Adding just the $L2$ loss for the cluster predictions.
    \item Adding just the $\mathcal{W}_2$ loss for the predicted posteriors.
    \item The final proposed loss function.
\end{itemize}
The results for this study are given in Table \ref{tab:ablation}. As it can be seen, our vanilla method itself performs decently. Adding just the scale-factor for STL-10 helps improve the accuracy. Just adding the $L2$ loss increases the accuracy by a small margin, while adding the $\mathcal{W}_2$ loss increases it by a larger margin. This shows that enforcing a consistent latent space representation helps make the learning more robust. Finally, we get the best performance by combining all our additions together.

\section{Conclusion and Future Works}
We propose an image clustering method using VAEs with a GMM prior where each component represents a cluster. The prior is learned jointly with the posterior, which in turn learns a strong latent representation that leads to accurate clustering, as shown in extensive experiments.
Our method doesn't require any pre-training and can be trained from scratch in and end-to-end manner.
We show results on a variety of datasets ranging from simple handwritten digits to complex real world objects, achieving state-of-the-art clustering accuracy among purely unsupervised methods. To the best of our knowledge, we are the first to achieve such significant results on real world datasets in a purely unsupervised manner. Moreover, we achieve comparable accuracy between similar datasets (CIFAR10/STL-10) in a transfer learning scenario. With the rise in the popularity of GANs, one way to move forward could be to replace the generator network in a GAN with our VAE model, thereby leading to a hybrid VAE-GAN model, that can has the strong latent representational powers of a VAE and the realistic generative powers of a GAN. Furthermore, adding explicit inference constraints on the GAN prior, somewhat as an increment along the lines of ClusterGAN\cite{mukherjee2019clustergan}, could be another way to approach the VAE-GAN hybrid architecture.

{\small
\bibliographystyle{IEEEtran}
\bibliography{egbib}
}

\end{document}